%% file: 00_main.tex
\documentclass[runningheads]{llncs}

\usepackage[T1]{fontenc}
\usepackage{graphicx, tabularx, graphics}
\usepackage{multirow}
\usepackage{booktabs}
\usepackage{url}
\usepackage{cite}
\usepackage{algpseudocode}
\usepackage{tabto}
\usepackage{arydshln}
\usepackage{algorithm}%
\usepackage[bottom, symbol]{footmisc}
\usepackage[hidelinks]{hyperref}
\usepackage{pifont}
\newcommand{\cmark}{\ding{51}}%
\newcommand{\xmark}{\ding{55}}%
\usepackage{amssymb}

\usepackage[dvipsnames]{xcolor}
\usepackage{color}

\begin{document}
\title{MEDBind: Unifying Language and Multimodal Medical Data Embeddings}
\titlerunning{MEDBind: Unifying Language and Multimodal Medical Data Embeddings}
\input{00_authors.tex}
\maketitle

\newcommand{\CM}[1]{{\footnote {\color{red}CM: #1}}}
\newcommand{\swk}[1]{{\footnote {\color{yellow}swk: #1}}}
\newcommand{\DA}[1]{{\footnote {\color{yellow}DA: #1}}}
\newcommand\figref{Fig.~\ref}

\begin{abstract}
\input{00_abstract.tex}
\end{abstract}

\input{01_introduction.tex}

\input{02_method.tex}

\input{03_results.tex}
\input{04_conclusion.tex}

\bibliographystyle{splncs04}
\bibliography{reference}

\input{05_appendix}

\end{document}

%% file: 00_authors.tex
\author{Yuan Gao\inst{1, 2, 4, 7}$^{,*}$ \and
Sangwook Kim\inst{1, 2, 7}$^{,*}$ \and
David E Austin\inst{1, 7} \and
Chris McIntosh\inst{1, 2, 3, 4, 5, 6, 7}}
\authorrunning{Y. Gao and S. Kim et al.}
\institute{
Peter Munk Cardiac Centre, University Health Network (UHN), Toronto, Canada \and
Department of Medical Biophysics, University of Toronto (U of T), Toronto, Canada \and
Department of Computer Science, U of T, Toronto, Canada \and
Ted Rogers Centre for Heart Research, UHN, Toronto, Canada \and
Toronto General Hospital Research Institute, UHN, Toronto, Canada \and
Department of Medical Imaging, U of T, Toronto, Canada \and
Vector Institute, Toronto, Canada \\
\email{\{yuan.gao, sangwook.kim\}@uhn.ca, deaustin@uwaterloo.ca, chris.mcintosh@uhn.ca} \\
}

\footnotetext[1]{Equal contribution.}

%% file: 00_abstract.tex
Medical vision-language pretraining models (VLPM) have achieved remarkable progress in fusing chest X-rays (CXR) with clinical texts, introducing image-text data binding approaches that enable zero-shot learning and downstream clinical tasks. However, the current landscape lacks the holistic integration of additional medical modalities, such as electrocardiograms (ECG). We present MEDBind (\textbf{M}edical \textbf{E}lectronic patient recor\textbf{D}), which learns joint embeddings across CXR, ECG, and medical text. Using text data as the central anchor, MEDBind features tri-modality binding, delivering competitive performance in top-K retrieval, zero-shot, and few-shot benchmarks against established VLPM, and the ability for CXR-to-ECG zero-shot classification and retrieval. This seamless integration is achieved through combination of contrastive loss on modality-text pairs with our proposed contrastive loss function, Edge-Modality Contrastive Loss, fostering a cohesive embedding space for CXR, ECG, and text. Finally, we demonstrate that MEDBind can improve downstream tasks by directly integrating CXR and ECG embeddings into a large-language model for multimodal prompt tuning.
The code for this project is available open public$^{\dagger}$.


\keywords{Vision-Language Pretraining \and Contrastive Learning \and Multimodal Deep Learning} \and Self-Supervised Learning

\footnotetext[2]{Code release under preparation at \url{https://github.com/mcintoshML/MedBind}}

%% file: 01_introduction.tex
\section{Introduction}
Vision-language pre-training models (VLPM) have advanced the integration of medical texts with imaging data, facilitating the convergence of diverse modalities into a unified representation space. This fusion deepens the understanding of text-image relationships and enhances their zero-shot learning capabilities.

VLPM have revolutionized the interpretation of chest X-ray images (CXR) by effectively aligning CXR with connected radiological reports through self-supervised contrastive learning. GloRIA\cite{huang2021gloria} and BioVIL\cite{boecking2022biovil} showcased the potential to discern local and global visual features in CXR through textual analysis. Further, MedCLIP\cite{wang2022medclip} and CXR-CLIP\cite{you2023_cxrclip} elevated training efficacy by improving image-text specific loss functions.
However, the scope of multimodal pre-training within the medical domain has been predominantly limited to image-text pairs, overlooking the potential integration of other clinical data types.

Incorporating more modalities from different domains is emerging as a critical research frontier. ImageBind\cite{girdhar2023imagebind} represents a significant stride in this direction by extending the VLPM contrastive learning approaches to accommodate more than two modalities within a unified embedding space, using images as the focal modality. ImageBind also broadened previous multimodal representations to additional tasks, including multimodal information retrieval and cross-modality zero-shot classification. Similarly, \textit{all in one}\cite{zhang2023allinone} achieved alignment by integrating video and text into a transformer for joint feature extraction across different modalities. Med-PaLM M\cite{tu2024towardsgoogle} recently advanced medical multimodal models by instruction prompt tuning PaLM-E, a large language model (LLM). Unlike contrastive learning approaches, Med-PaLM M incorporated multimodal data with text without explicit binding via LLM prompt tuning.

However, self-supervised contrastive learning in binding more than two medical modalities has yet to be explored. 
Thus, we introduce \textbf{MEDBind} (\textbf{M}edical \textbf{E}lectronic patient recor\textbf{D}), a contrastive learning model that explicitly binds CXR, electrocardiograms (ECG), and medical texts into a unified embedding space. We chose text as the central anchor for binding CXR and ECG since many medical modalities are interpreted and given clinical narrative summaries. 

\textbf{Contributions:} MEDBind is the first tri-modality framework that employs contrastive learning to fuse CXR, ECG, and medical texts into a unified representation space. We introduce a non-text edge-modality contrastive loss (EMCL) which strengthens the binding of CXR and ECG, and is adept at handling varying numbers of cross-modality pairs in datasets. MEDBind pretrained with EMCL improves information retrieval, zero-shot, and few-shot performance. We utilize MEDBind in downstream clinical tasks, where ECG and CXR embeddings are integrated with LLM to predict readmission and in-hospital mortality.

%% file: 02_method.tex
\section{Methods and Materials}
\input{02_01_method_modelframework}
\input{figures/figure2_framework}
\subsection{Loss function}


We trained MEDBind using Text-Modality Contrastive Loss (TMCL) for text-modality binding and Edge-Modality Contrastive Loss (EMCL), which is a novel loss function we propose for improving cross-modality binding.

CLIP\cite{radford2021learning_clip} showed that noise-contrastive estimation information (infoNCE) loss can bind image-text pairs, where a single positive-paired text is attracted for each image while the remaining texts are repelled. However, infoNCE loss does not account for cases where two patients have the same clinical text, incorrectly repelling their associated images. We implemented \textbf{TMCL},  similar to infoNCE, to link text with other modalities, but we considered identical paired texts as additional positive pairs. This is highlighted in the TMCL matrix of Fig. \ref{fig:framework}, where light grey pairs are additional related pairs with the same clinical text (for example, we encourage CXR with the same report to bind together). 

Adopting methods from \cite{khosla2020supervisedcontrastive, mo2024sclip}, we define TMCL in Eq. \ref{eq:TMCL}, where $z^j$ and $t^j$ denote embeddings for non-text modality and text, respectively (where $j\in \{c:CXR,\ e:ECG\}$). We denote $i, l \in n$ as the $i^{th}, l^{th}$ element in a batch size of $n$. 

\begin{equation}
    \label{eq:TMCL}
    L_{TMCL}^{t^j \to z^j} = -\sum_{i=1}^{n}\frac{1}{|P(i)|}
    \sum_{p \in P(i)}\log\frac{\exp(t^{j}_{i}\cdot z^j_p / \tau)}{\sum_{l=1}^{n}\exp(t^j_i\cdot z^j_l / \tau)}
\end{equation} \\where $p\in P(i)$ is the set of all positive pairs for text $t_{i}^j$ and modality $z^j$. The temperature parameter $\tau$ modulates the scale of distribution over embeddings. We use a symmetric loss for $L_{TMCL}$, so $L_{TMCL} = \sum_{j\in \{c, e\}}{}(L_{TMCL}^{t^j \to z^j} + L_{TMCL}^{z^j \to t^j})$.

We introduce \textbf{EMCL}, a novel contrastive loss that refines binding between non-text modalities. 
Unlike ImageBind \cite{girdhar2023imagebind}, EMCL explicitly binds CXR to ECG and can dynamically adapt to different CXR-ECG pair counts in a batch.  We defined positive pairs of CXR-ECG when a patient's CXR and ECG are taken during the same clinical visit, which pairs non-text modalities at the patient and temporal level.  We sub-sampled paired CXR-ECG instances from $n$ to optimize the usage for all training data. Thus, not all samples have corresponding CXR-ECG pairs (\figref{fig:framework}). Note that sub-sampling leads to a varying subset of size $m$ in each batch.
We define EMCL in Eq. \ref{eq:emcl}, where $u, q \in m$ is the $u^{th}, q^{th}$ element. 

\begin{equation}
    \label{eq:emcl}
    L_{EMCL}^{z^c \to z^e} = -\sum_{u=1}^{m}
    \log\frac{\exp(z^c_u\cdot z^e_u / \tau)}{\frac{n}{m}\sum_{q=1}^{m}\exp(z^c_u \cdot z^e_q / \tau)}
\end{equation}
 where the embeddings of CXR, $z^c$, and ECG, $z^e$, from the same patient case are aligned to a unified embedding space. EMCL stabilizes the fluctuating cardinality of this subset by normalizing the denominator with a factor of $\frac{n}{m}$ across different batch iterations. Similar to TMCL, we employed symmetric loss on $L_{EMCL}$ for bidirectional consistency, where  $L_{EMCL} = L_{EMCL}^{z_c \to z_e} + L_{EMCL}^{z_e \to z_c}$. 
 
 Our overall loss function is defined as $L_{TMCL}+L_{EMCL}$. Our $L_{EMCL}$ equips MEDBind to better navigate the complexity of cross-modality binding. 

\subsection{ECG-CLIP and Tri-modality Evaluations}
\label{subsection:ecgclip}


To our knowledge, no VLPM has bound ECG and text, making direct comparisons with existing models challenging.
Thus, we devised a novel \textbf{ECG-CLIP} as a baseline using our ECG and text encoders, trained with $L_{TMCL}$ where $j=e$. 

To assess the impact of tri-modality binding and EMCL, we introduce MEDBind;  \textbf{MEDBind\textsubscript{BD}} (bound) with $L_{TMCL} + L_{EMCL}$ and  \textbf{MEDBind\textsubscript{NM}} (normal) with only $L_{TMCL}$ as an ablation.
Moreover, we assessed if separately trained CXR and ECG VLPM could perform similarly to MEDBind, given that all VLPM bind the text modality. For tasks needing CXR and ECG encoders, we assessed various CXR VLPM paired with ECG-CLIP as the ECG encoder. This multiple single-paired VLPM approach is analogous to ``\textbf{encoder zoo}'' in \cite{moon2023anymal}.
\input{figures/figure1_tsne}
\input{tables/table_overview_v2}
\noindent\textbf{Implementation Details:} For training, we normalized CXR followed by augmentations\cite{wang2022medclip}. We normalized ECG and applied Gaussian noise augmentation. Input dimensions were \(224\times224\) for CXR and \(12\times1000\) for ECG. 
We truncated text to first 100 words without compromising the content. We set the final embedding size to 256 and temperature $\tau$ to 0.07. We trained models for 150 epochs with batch size 128, and used AdamW\cite{loshchilov2018decoupled_adamw} with weight decay 1e-1, learning rate 4e-4 adjusted via cosine annealing. We used PyTorch on an NVIDIA A100 GPU.

%% file: 02_01_method_modelframework.tex
\subsection{Model Architecture}

MEDBind is designed to process and analyze data from three distinct modalities: CXR, ECG, and medical text. Inspired by ImageBind\cite{girdhar2023imagebind}, our model employs dedicated encoders for each modality to extract representations (Fig. \ref{fig:framework}).


\textbf{Modality Encoder:} For the \textbf{CXR encoder}, we used Swin Transformer\cite{liu2021swin} as our backbone following \cite{wang2022medclip, you2023_cxrclip}. For the \textbf{ECG encoder}, we employed a vanilla transformer backbone \cite{vaswani2017attention}. We loaded ECG into a transformer by converting the time-series data into sequences of embeddings where each time point is tokenized using a linear embedding. We utilized BioBERT\cite{lee2020biobert}, a BERT\cite{devlin2018bert} variant fine-tuned on medical texts to capture biomedical semantics for the \textbf{Text encoder}. We opted not to apply BioClinicalBert\cite{alsentzer2019publicly_clinicalbiobert} for the text encoder to preserve the integrity of our training datasets for downstream tasks and to avoid potential bias since BioClinicalBert was fine-tuned on MIMIC-III. Finally, we used class token embeddings for all modality encoders because they are a critical component in transformer-based models that aggregate the global context of the input.
\textbf{Projection/Normalization}: We projected and normalized CXR, ECG, and text embeddings to 256 dimensions using modality-specific linear layer and L2 normalization. This ensures final embeddings are comparable across encoders.

%% file: figures/figure2_framework.tex
\begin{figure}[!ht]
    \centering
    \includegraphics[width=\textwidth]{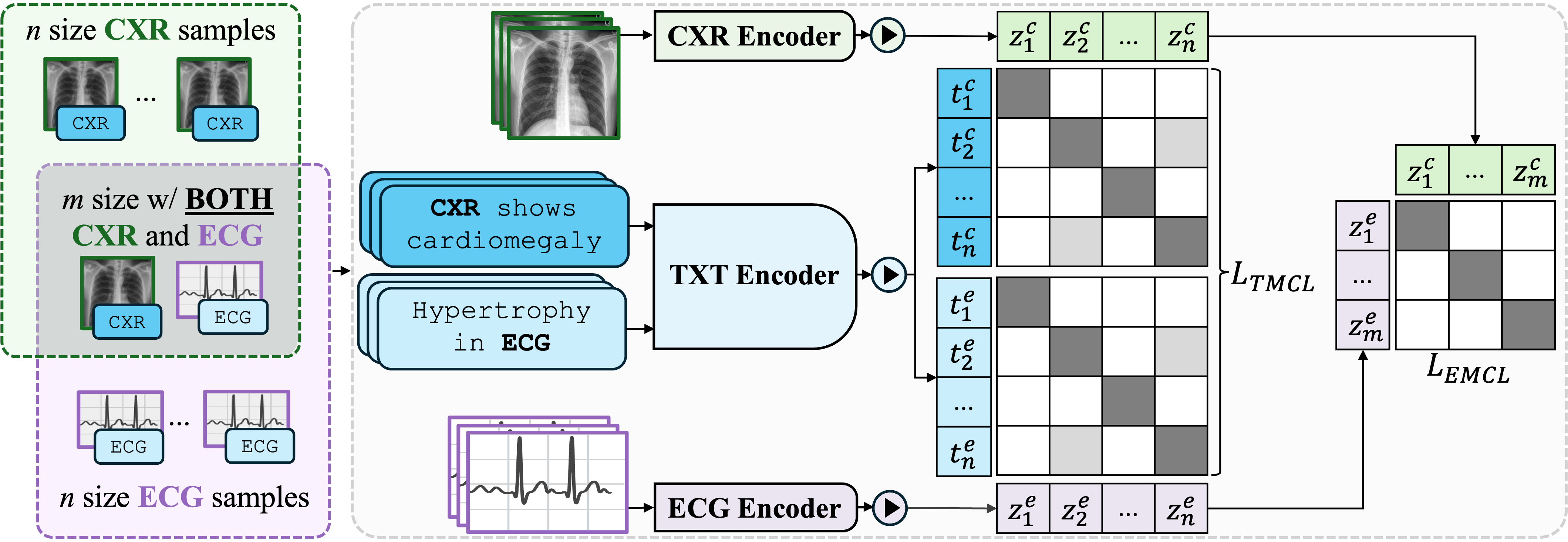}
    \caption{Proposed method. Batch size $n$: CXR (green), ECG (purple), and paired text (blue). Subset size $m$: paired ECG/CXR. Inputs are embedded and normalized ($\blacktriangleright$). We used two losses: 1) Text-Modality Contrastive Loss (TMCL); 2) Edge-Modality Contrastive Loss (EMCL). Grey is positive-pair; light grey is additional related pairs.}
    \label{fig:framework}
\end{figure}

%% file: figures/figure1_tsne.tex
\begin{figure}[ht!]
    \centering
    \includegraphics[width=\textwidth]{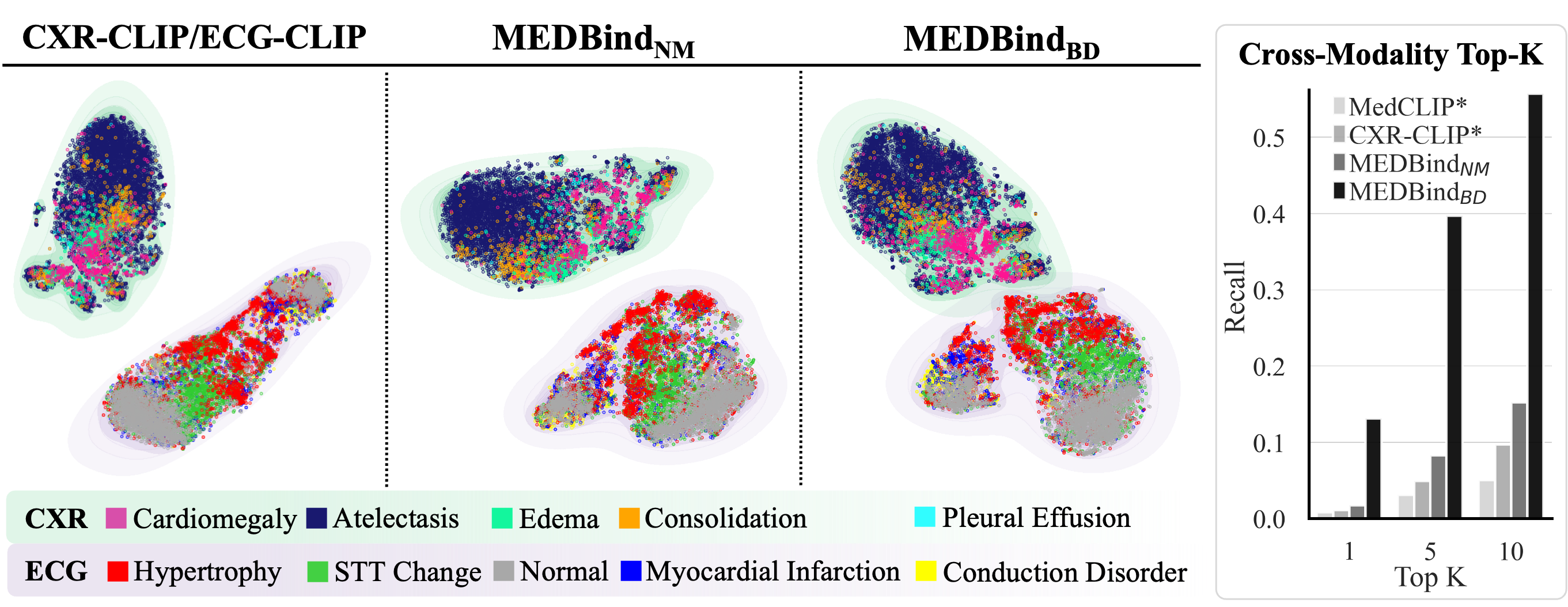}
    \caption{Embedding visualization and CXR-to-ECG cross-modality retrieval. (Left) t-SNE plots of CXR and ECG embeddings for various models. (Right) Cross-modality retrieval Top-K recall. MEDBind\textsubscript{BD} brings CXR and ECG clusters closer in t-SNE and tops cross-modality recall@\{1,5,10\}. $^*$CXR VLPM with ECG-CLIP as encoder zoo.}
    \label{fig:tsne}
\end{figure}

%% file: tables/table_overview_v2.tex
\begin{table}[ht!]
    \label{tab:dataset}
  \centering
  \caption{Overview of datasets, tasks, class count (CLS), and training split. LINK shows connected datasets. $^*$Subset from MIMIC-CXR and ECG with both CXR/ECG.}
  
 {
     \setlength{\dashlinedash}{0.3pt}
    \setlength{\dashlinegap}{1.5pt}
  \def\thickhline{\noalign{\hrule height1.0pt}}
    \begin{tabular}{llccccc}

    \thickhline
    Dataset & Task &LINK & CLS & Train & Valid  & Test \\
    \hline
    MIMIC-CXR\cite{johnson2019mimiccxr} & Pretrain/Retrieval/LLM-Prompt & \cmark & 12 & 86,853 & 12,059 & 24,799 \\
    Open-I\cite{demner2016preparing_openi} & Retrieval & \xmark & - & - & - & 3,269 \\
    CheXpert\cite{irvin2019chexpert} & Retrieval & \xmark & 5 & - & - & 1,000 \\
    COVID\cite{chowdhury2020can_kaggle_covid1} & Zero-Shot/Few-Shot & \xmark & 2 & 11,028 & - & 2,780 \\
    RSNA\cite{shih2019augmenting_rsna} & Zero-Shot/Few-Shot & \xmark & 2 & 18,678 & - & 5,338 \\
    \hdashline[0.3pt/1.5pt]
    MIMIC-ECG\cite{gow_mimicecg} & Pretrain/Retrieval/LLM-Prompt & \cmark & 5 & 88,291 & 12,065 & 24,644 \\
    PTB-XL\cite{strodthoff2020deep_ptbxl}  & Retrieval/Zero-Shot/Few-Shot & \xmark & 5 & 17,415 & 2,183 & 2,198 \\
    ICBEB\cite{liu2018open_icbeb} & Zero-Shot/Few-Shot & \xmark  & 9 & 5,501 & - & 1,376 \\
    \hdashline[0.3pt/1.5pt]
    MIMIC-IV\cite{johnson2023mimic4} & LLM-Prompt & \cmark  & - & 218,787 & 40,995 & 72,473 \\
    MIMIC-PAIR* & Pretrain/Retrieval/LLM-Prompt & \cmark  & - & 22,397 & 3,292 & 6,664 \\  
    \thickhline
    
    \end{tabular} %
    }
    \label{tab:datasets}
\end{table}

%% file: 03_results.tex
\section{Experiments and Results}
\input{03_01_datasets}
\subsection{Modality-to-Text and Cross-Modality Retrieval}
\textbf{Modality-to-Text Retrieval:} MEDBind preserved retrieval integrity for CXR and ECG compared to single-paired VLPM. We tested CXR and ECG \textit{modality}-to-text retrieval accuracy using recall for the top-K correct clinical reports. In Table \ref{tab:performanceMetrics}, we benchmarked MEDBind against MedCLIP\cite{wang2022medclip} and CXR-CLIP\textsubscript{\textit{SwinT}} \cite{you2023_cxrclip} for CXR-to-text retrieval, and ECG-CLIP for ECG-to-text retrieval. MEDBind outperformed all separately trained VLPM in total RSUM \textit{modality}-to-text retrieval. However, in Open-I dataset, MedBind\textsubscript{NM} outperformed MedBind\textsubscript{BD}, indicating that binding may lose some task depth in exchange for its breadth. \\
\textbf{Cross-Modality Retrieval:} In addition to \textit{modality}-to-text retrieval, we also evaluated cross-modality retrieval between CXR and ECG on MIMIC-PAIR test set (Section \ref{sec:datasets}) and compared to encoder zoo (CXR-CLIP or MedCLIP with ECG-CLIP). In \figref{fig:tsne}, top-K recall for cross-modality retrieval highlights MEDBind\textsubscript{BD}, with EMCL, outperforms MEDBind\textsubscript{NM} and other models.
We visualize t-SNE plots to qualitatively prove that MEDBind\textsubscript{BD} brings CXR and ECG clusters closer within a joint space while maintaining class clustering, compared to other models. These results demonstrate the ability of MEDBind\textsubscript{BD} to match CXR to ECG and project modalities within a unified space.

\input{tables/table_modalitytotext}
\input{figures/figure3_fewshot}
\input{tables/table_xmodality_zs}
\subsection{Zero/Few-shot and Cross-Modality Classification}
\textbf{Modality-to-Text Zero/Few-shot:} 
We evaluated zero-shot performance by calculating the cosine distance between text embeddings and non-text modality embeddings following \cite{wang2022medclip}. We performed few-shot classification using embeddings from the frozen CXR or ECG encoders via the linear probing method \cite{girdhar2023imagebind}. We reported the average balanced accuracy for each shot over 300 different support sets. In \figref{fig:fewshot}, we compared MEDBind with other state-of-the-art models. In zero-shot, MEDBind\textsubscript{BD} consistently beat MEDBind\textsubscript{NM} across all datasets and outperformed other models in three out of four datasets. MEDBind\textsubscript{BD} also maintained strong performance in all few-shot scenarios. Results show that EMCL boosted performance without compromising CXR or ECG zero and few-shot capabilities. Notably, MEDBind\textsubscript{BD}'s zero-shot exceeded few-shot performance in the COVID dataset, highlighting its robustness on unseen classes.

\input{tables/table_downstream_tasks}
\noindent\textbf{Cross-Modality Zero-Shot}: Cardiomegaly and hypertrophy are commonly diagnosed from CXR and ECG, respectively. Both diseases can manifest pathophysiological signs detectable in CXR and ECG\cite{hypertrophy}. Thus, we introduce a novel cross-modality zero-shot classification task, assessing if we can detect hypertrophy via CXR and cardiomegaly via ECG, on MIMIC-PAIR test set.
We calculated cosine distances between query and support embeddings. For example, we used CXR as query with ECG as support to predict hypertrophy.
In Table \ref{tab:crossmodality}, results showed that MEDBind\textsubscript{BD} outperformed MEDBind\textsubscript{NM} and encoder zoo. MEDBind\textsubscript{BD}'s strong cross-modality zero-shot performance implies its ability to integrate CXR and ECG into a unified space--a unique advantage of our EMCL.

\subsection{Multimodal LLM Integration}
To assess the efficacy of MEDBind in integrating cross-modality data with free text into LLM, we conducted experiments on predicting 30-day hospital readmission and in-hospital mortality\cite{jiang2023health_nature}. We used BioBERT as LLM due to its compatibility with BERT-based VLPM. Using MIMIC-IV, we provided BioBERT with discharge summaries for readmission and patient demographics notes for mortality predictions (Appendix). We excluded discharge texts to prevent bias in mortality information. If CXR or ECG were connected to a clinical visit, we provided the following inputs after the text: 1) CXR and ECG clinical interpretation from experts (\textbf{Text-only}), 2) embeddings from \textbf{encoder zoo}, or 3) embeddings from \textbf{MEDBind}.  We extracted embeddings from frozen non-text modality encoders with a trainable linear projection layer for LLM integration following \cite{moon2023anymal}. We used Low-Rank Adaption for the efficient LLM prompt-tuning\cite{hu2021lora}.   

Table \ref{tab:downstream} highlights the performance of MEDBind\textsubscript{BD} for prompt tuning LLM, BioBERT, compared to MEDBind\textsubscript{NM} and encoder zoo. MEDBind\textsubscript{BD} outperforms its counterparts by binding CXR and ECG pairs using our proposed EMCL. While the text-only LLM performs similarly on downstream tasks, it relies on clinician-generated texts. Instead, MEDBind\textsubscript{BD} is more automated as it can directly process CXR and ECG--increasing clinical workflow efficiency. 

 

%% file: 03_01_datasets.tex
\subsection{\textbf{Datasets}}\label{sec:datasets}
We present datasets and details in Table \ref{tab:datasets}. We pretrained MEDBind on MIMIC-CXR and MIMIC-ECG, including MIMIC-PAIR subset. To avoid training contamination, we maintained the same patient-level splits for all MIMIC datasets.

Starting with CXR datasets, \textbf{MIMIC-CXR}\cite{johnson2019mimiccxr} consists of CXR with their paired reports and labels\cite{PhysioNet}. We pre-processed CXR and text using methods from MedCLIP\cite{wang2022medclip}. In this study, we only included AP and PA view CXR.
\textbf{CheXpert}\cite{irvin2019chexpert} consists of a large number of CXR. Like \cite{wang2022medclip, you2023_cxrclip}, we formed \textbf{CheXpert5x200} with 200 randomly selected CXR from 5 classes in \cite{irvin2019chexpert}. We generated prompts for CXR-text retrieval tasks, as proposed in \cite{wang2022medclip}. 
\textbf{COVID}\cite{chowdhury2020can_kaggle_covid1} is a public dataset with binary COVID-19 labels. We generated prompts as suggested in \cite{wang2022medclip} (details in Appendix). 
\textbf{RSNA}\cite{shih2019augmenting_rsna} contains pneumonia cases from CXR, publicly available in the National Institutes of Health database.

\textbf{MIMIC-ECG}\cite{gow_mimicecg} has 10-second 12-lead ECG at 500Hz, downsampled to 100Hz using a low-pass filter\cite{kher2019signalecg}. ECG has machine reports and links (\texttt{cart\_id}) to free-form text. We used free-form text where available or machine reports to generate ECG text. We created labels (Hypertrophy, STT Change, Normal, Myocardial Infarction, Conduction Disorder) from text using a rule-based method inspired by \cite{irvin2019chexpert}, excluding ECG with undetectable labels (Appendix).
\textbf{PTB-XL}\cite{strodthoff2020deep_ptbxl} has 10 second 12-lead ECG at 100 Hz, and superclass labels \cite{PhysioNet,strodthoff2020deep_ptbxl}. 
\textbf{ICBEB}\cite{liu2018open_icbeb} has 6-60 seconds 12 leads ECG at 100 Hz and class labels. We used the first 10 seconds for all ECG and zero-padded shorter ECG to 10 seconds. 

\textbf{MIMIC-IV}\cite{johnson2023mimic4} contains health records from patients in MIMIC, including discharge notes. We derived in-hospital mortality labels with \texttt{discharge\_loc} and 30-day readmission from if a patient had a subsequent visit within a 30-day window, using patient's \texttt{subject\_id} and admission time \texttt{admittime}. 
We linked CXR and ECG to MIMIC-IV by \texttt{subject\_id} and modality recording times within 24 hours. Using this pairing strategy, we linked MIMIC-CXR and MIMIC-ECG, referred to as \textbf{MIMIC-PAIR}, by linking visit identifiers (\texttt{hadm\_id}) in MIMIC-CXR and MIMIC-ECG if available. Without \texttt{hadm\_id}, we paired cases on \texttt{subject\_id} and if CXR and ECG recording times were within 24 hours.

%% file: tables/table_modalitytotext.tex
\def\thickhline{\noalign{\hrule height1.0pt}}
\begin{table}[ht!]
\centering
\caption{Results of CXR and ECG \textit{modality}-to-text retrieval. Recall@K =\{1,10\} (R\textsubscript{K}). C5x200 is CheXpert5x200. Total RSUM is sum of R\textsubscript{K} per modality. \textbf{Bold}=best, \underline{underline}=Second best. $^*$data splits differed from CXR-CLIP, so results taken from \cite{you2023_cxrclip}. }
\label{tab:performanceMetrics}

\begin{tabular}{c;{0.3pt/1.5pt}cc;{0.3pt/1.5pt}cc;{0.3pt/1.5pt}cc;{0.3pt/1.5pt}cc;{0.3pt/1.5pt}cc|cc}

\thickhline
\multirow{2}{*}{Model} & \multicolumn{2}{c;{0.3pt/1.5pt}}{MIMIC-CXR} & \multicolumn{2}{c;{0.3pt/1.5pt}}{C5x200} & \multicolumn{2}{c;{0.3pt/1.5pt}}{Open-I} & \multicolumn{2}{c;{0.3pt/1.5pt}}{MIMIC-ECG} & \multicolumn{2}{c|}{PTB-XL} &\multicolumn{2}{c}{Total RSUM} \\ 

                      & R\textsubscript{1} & R\textsubscript{10} & R\textsubscript{1} & R\textsubscript{10} & R\textsubscript{1} & R\textsubscript{10} & R\textsubscript{1} & R\textsubscript{10} & R\textsubscript{1} & R\textsubscript{10} & CXR & ECG \\ \hline

CLIP                   & 1.0 & 10.5 & 1.1 & 13.9 & 0.8 & 8.1 & - & - & - & - & 35.4 & -\\
MedCLIP                & 2.8 & 18.0 & \textbf{2.9}&  \textbf{31.2} & 0.9  & 8.6 & - & - & - & - & 64.4 & -\\
CXR-CLIP              &  $21.6^*$ & $60.2^*$& 2.2  & 20.3 & \underline{14.1}  & 39.3 & - & - & - & - & 157.7 & -\\
ECG-CLIP               & - & - & - & - & - & - & \underline{51.5} & \textbf{95.5} & \textbf{2.1} & 17.4 & - & \underline{166.5} \\
\hline
MEDBind\textsubscript{NM} & \underline{43.8} & \underline{88.5} & \underline{2.4} & \underline{22.1} & \textbf{14.3} & \textbf{41.1} & 50.2 & 93.9 & \underline{1.9} & \underline{18.2} & \textbf{212.2} & 164.2\\
MEDBind\textsubscript{BD}  & \textbf{44.7} & \textbf{91.0} & \underline{2.4} & 20.0 & 13.6 &  \underline{39.7} & \textbf{53.6}  & \underline{94.5} & 1.6  & \textbf{19.2} & \underline{211.4} & \textbf{168.9} \\ 
\thickhline
\end{tabular}
\end{table}

%% file: figures/figure3_fewshot.tex
\begin{figure}[!ht]
    \centering
    \includegraphics[width=\textwidth]{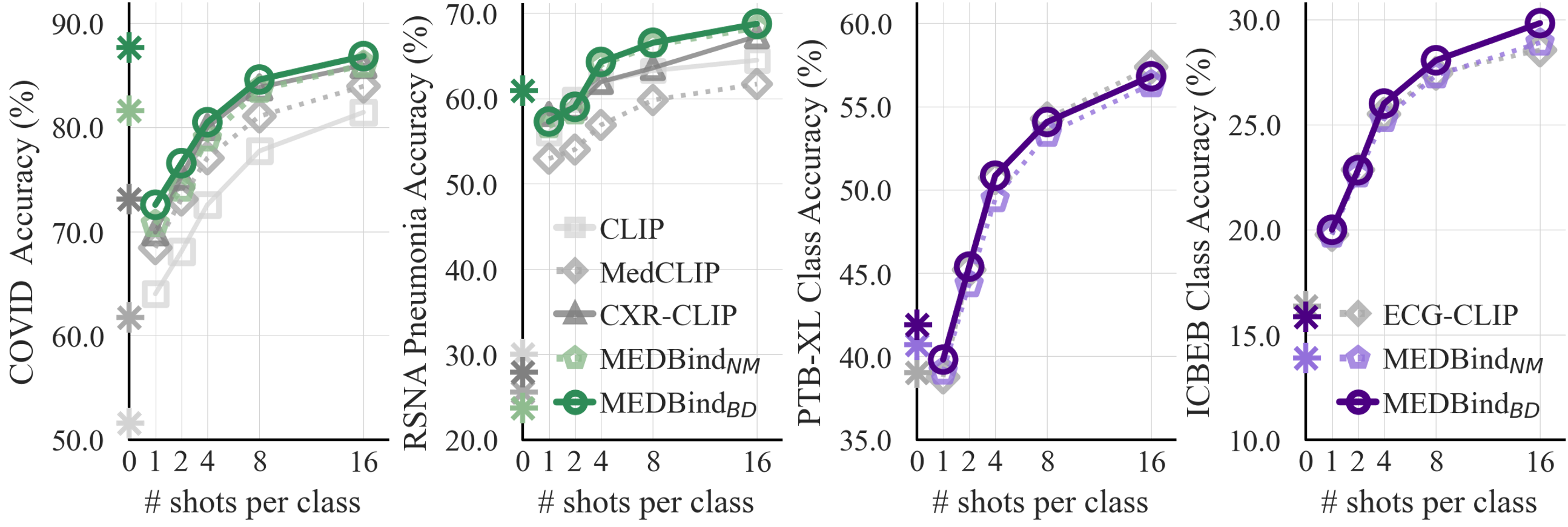}
    \caption{Results of zero-shot (denoted as astericks (*) on y-axis) and few-shot (K=\{1,2,4,8,16\}) classification using balanced accuracy (\%) on CXR (green): COVID and RSNA datasets, and ECG (purple): PTB-XL and ICBEB datasets. 
    }
    \label{fig:fewshot}
\end{figure}

%% file: tables/table_xmodality_zs.tex
\begin{table}[ht!]
\def\thickhline{\noalign{\hrule height1.0pt}}

\centering
\caption{Cross-modality zero-shot performance, ACC refers to balanced accuracy (\%).}
\setlength{\tabcolsep}{5pt}
\begin{tabular}{c;{0.3pt/1.5pt}c;{0.3pt/1.5pt}c;{0.3pt/1.5pt}c|c}

    \thickhline
    Cross-Modality Zero-shot Task & Inputs & Query & Support & ACC \\
    \hline

    \multirow{4}{*}{\parbox{4.5cm}{\centering \textbf{Hypertrophy vs. Other} \\ \textit{Given CXR (query), \\predict its ECG class using\\ ECG support set}}} & \multirow{4}{*}{CXR}
    & MedCLIP & ECG-CLIP & 60.9 \\
    & & CXR-CLIP & ECG-CLIP & 72.9 \\
    & & MEDBind\textsubscript{NM} & MEDBind\textsubscript{NM} & \underline{73.7} \\
    & & MEDBind\textsubscript{BD}& MEDBind\textsubscript{BD} & \textbf{82.1} \\
    \hline
    \multirow{4}{*}{\parbox{4.5cm}{\centering \textbf{Cardiomegaly vs. Other} \\ \textit{Given ECG (query), \\predict its CXR class using CXR support set}}} & \multirow{4}{*}{ECG} &  ECG-CLIP & MedCLIP & \underline{73.5} \\
    & & ECG-CLIP & CXR-CLIP & 70.2 \\
    & & MEDBind\textsubscript{NM} & MEDBind\textsubscript{NM} & 69.1 \\
    & & MEDBind\textsubscript{BD} & MEDBind\textsubscript{BD} & \textbf{84.6} \\
    \thickhline
\end{tabular} %
    
  \label{tab:crossmodality}
\end{table}

%% file: tables/table_downstream_tasks.tex
\begin{table}[ht!]
    \centering
    
    \caption{LLM prompt tuning task. Inputs include: medical text (TXT\textsubscript{MD}); CXR/ECG text (TXT\textsubscript{C}/ TXT\textsubscript{E}); CXR/ECG embedding (EMB\textsubscript{C}/EMB\textsubscript{E}).  $^*$discharge and admission text used for 30-day readmission (Readmit.) and in-hospital mortality (In Hosp.). \textbf{Bold} and \underline{underline} denote best and second best mixed input models (\cmark), respectively.}
    \def\thickhline{\noalign{\hrule height1.0pt}}
    
    \begin{tabular}{cccc;{0.3pt/1.5pt}c;{0.3pt/1.5pt}c}
        \thickhline
        \multirow{2}{*}{Method}  & \multirow{2}{*}{\parbox{1cm}{\centering Mixed\\ Input}} & \multirow{2}{*}{\parbox{2.7cm}{\centering CXR/ECG \\ Interpreter}} & \multirow{2}{*}{LLM Inputs} & \multicolumn{1}{c;{0.3pt/1.5pt}}{Readmit.} & \multicolumn{1}{c}{In Hosp.} \\
        & & & & ACC &  ACC \\
        
        \hline
        Text-Only & \xmark & \textit{Clinical Expert} & TXT\textsubscript{MD$^{*}$,C,E} &65.0 & 74.5 \\
        \hline
        \multirow{2}{*}{Encoder zoo} & \cmark & MedCLIP/ECG-CLIP & \multirow{2}{*}{ TXT\textsubscript{MD}$^{*}$+EMB\textsubscript{C,E}}  & 60.5 & 72.0 \\
        & \cmark & CXR-CLIP/ECG-CLIP & & 59.9 & 71.6 \\
        \hdashline[0.3pt/1.5pt]
        \multirow{2}{*}{MEDBind} & \cmark & MEDBind\textsubscript{NM} & \multirow{2}{*}{ TXT\textsubscript{MD}$^{*}$+EMB\textsubscript{C,E}} & \underline{60.5} & \underline{73.6} \\
        & \cmark & MEDBind\textsubscript{BD} & & \textbf{64.3}& \textbf{74.8} \\
        \thickhline
    \end{tabular} %
       \label{tab:downstream}
\end{table}

%% file: 04_conclusion.tex
\section{Conclusion}\label{sec:conclusion}
We introduced MEDBind, a tri-modality binding framework integrating multimodal medical data of CXR, ECG, and text. We demonstrated its benefits in binding different modalities into a unified space via EMCL, which enhanced zero-shot and downstream task performance over single-paired VLPM. Our method is scalable and open for future expansion to include additional modalities.

%% file: 05_appendix.tex
\clearpage
\appendix
\setcounter{figure}{0} \renewcommand{\thefigure}{A.\arabic{figure}}
\setcounter{table}{0} \renewcommand{\thetable}{A.\arabic{table}}
\setcounter{page}{1}
\section{Appendix}
\def\thickhline{\noalign{\hrule height1.0pt}}
\begin{table}[ht!]
\small
\caption{Labels and prompts used for zero-shot evaluation for each dataset. We denote \texttt{\{LABEL\}} as the label associated with the patient case. For PTB-XL and ICBEB, \texttt{\{LABEL\}} were used to generate prompts. Since there are no available clinical reports for COVID dataset, we generated text prompts similar to MedCLIP\cite{wang2022medclip} to include some radiological findings. We randomly selected 10 out of 20 generated prompts per label for COVID dataset and calculated the average cosine distance in zero-shot settings.}
\begin{tabular}{l;{0.3pt/1.5pt}l;{0.3pt/1.5pt}l}

\thickhline
Dataset & Labels & Prompts \\ \hline
\multirow{7}{1.25cm}{COVID} & \multirow{7}{5cm}{COVID-19, Normal} & \parbox[t]{5.4cm}{\textbf{COVID-19}:[\textit{``Multifocal bilateral opacities.'', ``Atypical pneumonia with peripheral distribution and sparing of the lung apices.'' ...}] \\ \textbf{Normal}:[\textit{``Heart size is normal and the lungs are clear.'', ``The heart is normal in size and contour.'', ...}]} \\
\cline{1-1} \cline{2-2} \cline{3-3}
\multirow{3}{1.25cm}{RSNA} & \multirow{3}{5cm}{Pneumonia, Normal} & \parbox[t]{5.4cm}{\textbf{Pneumonia}:\textit{``Findings suggesting \\ pneumonia.''} \\ \textbf{Normal}:``\textit{No evidence of pneumonia.}''} \\
\cline{1-1} \cline{2-2} \cline{3-3}
\multirow{3}{1.25cm}{PTB-XL} & \parbox[t]{5cm}{Hypertrophy, Myocardial Infarction, STT Changes, Conduction disturbance, Normal sinus rhythm} & \multirow{3}{5.4cm}{``\textit{This ECG shows} \texttt{\{LABEL\}}.''} \\
\cline{1-1} \cline{2-2} \cline{3-3}
\multirow{6}{1.25cm}{ICBEB} & {\parbox[t]{5cm}{First-degree atrioventricular block, atrial fibrillation, complete left/right bundle branch block, normal sinus rhythm, premature atrial contraction, ST-segment depression, ST-segment elevated}} & \multirow{6}{5.4cm}{``\textit{This ECG shows} \texttt{\{LABEL\}}.''} \\
\bottomrule
\end{tabular}
\label{tab:appendixprompt}
\end{table}
\begin{table}[ht!]
\small
\caption{Generated text for MIMIC-ECG and MIMIC-IV. Inputs are incorporated into the \textbf{Generated Text Format}.
Each ECG contains a list of machine reports (i.e. \texttt{report\_0}) for each ECG. MIMIC-IV-generated texts were used for in-hospital mortality.}
\begin{tabular}{l;{0.3pt/1.5pt}l;{0.3pt/1.5pt}l}
\thickhline
Datasets & Inputs & Generated Text Format \\ \hline
\multirow{3}{2cm}{MIMIC-ECG} & \multirow{3}{4.3cm}{\texttt{[report\_0, report\_1,\\…, report\_17]}} & \parbox[t]{5.5cm}{\textit{``}\textit{ECG presents }\texttt{\{report\_0\}}.\textit{ Additional findings include the following:} \texttt{\{report\_1, ..., report\_17\}}.\textit{''}} \\
\hline
\multirow{4}{2cm}{MIMIC-IV} & \multirow{4}{4.3cm}{\texttt{[gender, anchor\_age, admission\_type, admission\_location]}} & \parbox[t]{5.5cm}{\textit{``}\texttt{\{gender\}} \textit{patient, who is at the age of} \texttt{\{anchor\_age\}}\textit{, was admitted as }\texttt{\{admission\_type\}}\textit{. Location:} \texttt{\{admission\_location\}.}\textit{''}
} \\
\bottomrule
\end{tabular}
\label{tab:appendixrulebased}
\end{table}
\input{figures/figure4_prompttuning}
\def\thickhline{\noalign{\hrule height1.0pt}}
\begin{table}[ht!]
\caption{Rule-based method to label MIMIC-ECG data using clinical text. We label each class based on the expert-generated keywords found in \cite{strodthoff2020deep_ptbxl} (\textbf{Keywords}) and excluded ECG if the associated clinical text contained content presented in \textbf{Disallowed Content}, which represents poor quality data.}
\begin{tabular}{l;{0.3pt/1.5pt}l;{0.3pt/1.5pt}l}
\thickhline
Class & Disallowed Content & Keywords \\ \hline
\multirow{3}{2cm}{Normal (NORM)} & \multirow{18}{4cm}{\texttt{[borderline ecg, \\ poor quality, without \\ knowing patient, error, pediatric, warning: data quality, missing lead, unsuitable for analysis, motion artifacts, requires manual review, technical difficulties, possibly, probable]}} & \parbox[t]{5.5cm}{\texttt{[normal ecg, no issues found, normal ekg, normal heart tracing, within normal limits]}} \\
\cline{1-1} \cline{3-3}
\multirow{3}{2cm}{Hypertrophy (HYP)} &  & \parbox[t]{5.5cm}{\texttt{[hypertrophy, left atrial enlargement, LVH, LAO, overload, enlargement]}} \\
\cline{1-1} \cline{3-3}
\multirow{5}{2cm}{STT Changes (STTC)} &  & \parbox[t]{5.5cm}{\texttt{[ST elevation, T wave changes, nonspecific T abnormalities, ST changes, T changes, ventricular premature complex, VPC, PVC, ST change]}} \\
\cline{1-1} \cline{3-3}
\multirow{3}{2.2cm}{Myocardiac Infarction (MI)} &  & \parbox[t]{5.5cm}{\texttt{[myocardial ischemia, inferior infarct, anterior infarct, septal infarct]}} \\
\cline{1-1} \cline{3-3}
\multirow{5}{2cm}{Conduction Disorder (CD)} &  & \parbox[t]{5.5cm}{\texttt{[degree A-V block, PAC, prolonged PR interval, conduction delay, left axis deviation, bundle branch block, pacemaker, atrial pacing, rBB, LAFB, PVC]}} \\
\bottomrule
\end{tabular}
\label{tab:ecgmimic_textcreation}
\end{table}
\begin{table}[ht!]
\newcolumntype{C}{>{\centering\arraybackslash}X}
\centering
\caption{MIMIC-ECG class distribution using our rule-based approach. The table highlights the number of ECG cases in MIMIC-ECG detected. $^*$N/A column represents ECGs that our approach could not label and were excluded from our study.}
\begin{tabularx}{\textwidth}{C;{0.3pt/1.5pt}C;{0.3pt/1.5pt}C;{0.3pt/1.5pt}C;{0.3pt/1.5pt}C;{0.3pt/1.5pt}C} 
\thickhline 
NORM & HYP & STTC & MI & CD & N/A* \\ 
\hline
34,097 & 11,305 & 28,918 & 19,242 & 31,438 & 670,939 \\
\bottomrule
\end{tabularx}
\label{tab:ecg_classdistribution}
\end{table}


%% file: figures/figure4_prompttuning.tex
\begin{figure}[!ht]
    \centering
    \includegraphics[width=\textwidth]{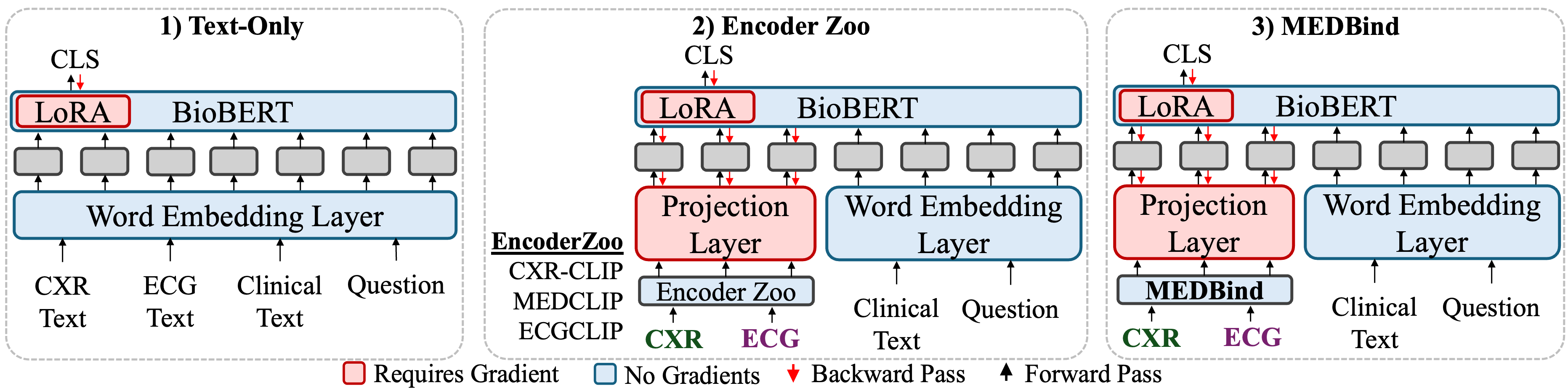}
    \caption{Three different training paradigms for downstream LLM tasks. 1) Text-only: Traditional method of prompt tuning using LoRA\cite{hu2021lora} to tune weights of BioBERT\cite{lee2020biobert}. 2) Encoder Zoo: AnyMAL\cite{moon2023anymal} paradigm for fine-tuning, which incorporates multiple modalities by inputting CXR and ECG tokens—generated either from CXR-CLIP\cite{you2023_cxrclip} and ECG-CLIP or MedCLIP\cite{wang2022medclip} and ECG-CLIP alongside clinical text. 3) MEDBind: which is a unified model for multimodal binding.}
    \label{fig:prompttuning}
\end{figure}